# Kinematic Analysis of a Serial – Parallel Machine Tool: the VERNE machine

Daniel Kanaan, Philippe Wenger and Damien Chablat

Institut de Recherche en Communications et Cybernétique de Nantes UMR CNRS 6597

1, rue de la Noë, BP 92101, 44312 Nantes Cedex 03 France

E-mail address: Daniel.Kanaan@irccyn.ec-nantes.fr

**Abstract**

*The paper derives the inverse and the forward kinematic equations of a serial – parallel 5-axis machine tool: the VERNE machine. This machine is composed of a three-degree-of-freedom (DOF) parallel module and a two-DOF serial tilting table. The parallel module consists of a moving platform that is connected to a fixed base by three non-identical legs. These legs are connected in a way that the combined effects of the three legs lead to an over-constrained mechanism with complex motion. This motion is defined as a simultaneous combination of rotation and translation. In this paper we propose symbolical methods that able to calculate all kinematic solutions and identify the acceptable one by adding analytical constraint on the disposition of legs of the parallel module.*

## 1. Introduction

Parallel kinematic machines (PKM) are well known for their high structural rigidity, better payload-to-weight ratio, high dynamic performances and high accuracy [1, 2, 3]. Thus, they are prudently considered as attractive alternatives designs for demanding tasks such as high-speed machining [4]. Most of the existing PKM can be classified into two main families. The PKM of the first family have fixed foot points and variable–length struts, while the PKM of the second family have fixed length struts with moveable foot points gliding on fixed linear joints [5, 6].

In the first family, we distinguish between PKM with six degrees of freedom generally called Hexapods and PKM with three degrees of freedom called Tripods [7, 8]. Hexapods have a Stewart–Gough parallel kinematic architecture. Many prototypes and commercial hexapod PKM already exist, including the VARIAX (Gidding and Lewis), the TORNADO 2000 (Hexel). We can also find hybrid architectures such as the TRICEPT machine (SMT Tricept) [9], which is composed of a two-axis wrist mounted in series to a 3-DOF "tripod" positioning structure.

In the second family, we find the HEXAGLIDE (ETH Zürich) that features six parallel and coplanar linear joints. The HexaM (Toyoda) is another example with three pairs of adjacent linear joints lying on a vertical cone [10]. A hybrid parallel/kinematic PKM with three inclined linear joints and a two-axis wrist is the GEORGE V (IFW Uni Hanover).

Many three-axis translational PKMs belong to this second family and use architecture close to the linear Delta robot originally designed by Clavel for pick-and-place operations [11]. The Urane SX (Renault Automation) and the QUICKSTEP (Krause and Mauser) have three non-coplanar horizontal linear joints [12].

Because many industrial tasks require less than six degrees of freedom, several lower-DOF PKMs have been developed [13-15]. For some of these PKMs, the reduction of the number of DOFs can result in coupled motions of the mobile platform. This is the case, for example, in the RPS manipulator [13] and in the parallel module of the Verne machine. The kinematic modeling of these PKMs must be done case by case according to their structure.

Many researchers have contributed to the study of the kinematics of lower-DOF PKMs. Many of them have focused on the discussion of both analytical and numerical methods [16, 17]. This paper investigates the inverse and direct kinematics of the VERNE machine and derives closed form solutions. The VERNE machine is a 5-axis machine-tool that was designed by Fatronik for IRCCyN [18, 19]. This machine-tool consists of a parallel module and a tilting table as shown in Fig. 1. The parallel module moves the spindle mostly in translation while the tilting table is used to rotate the workpiece about two orthogonal axes.

The purpose of this paper is to formulate analytic expressions in order to find all possible solutions for the inverse and forward kinematics problem of the VERNE machine. Then we identify and sort these solutions in order to find the one that satisfies the end-user.

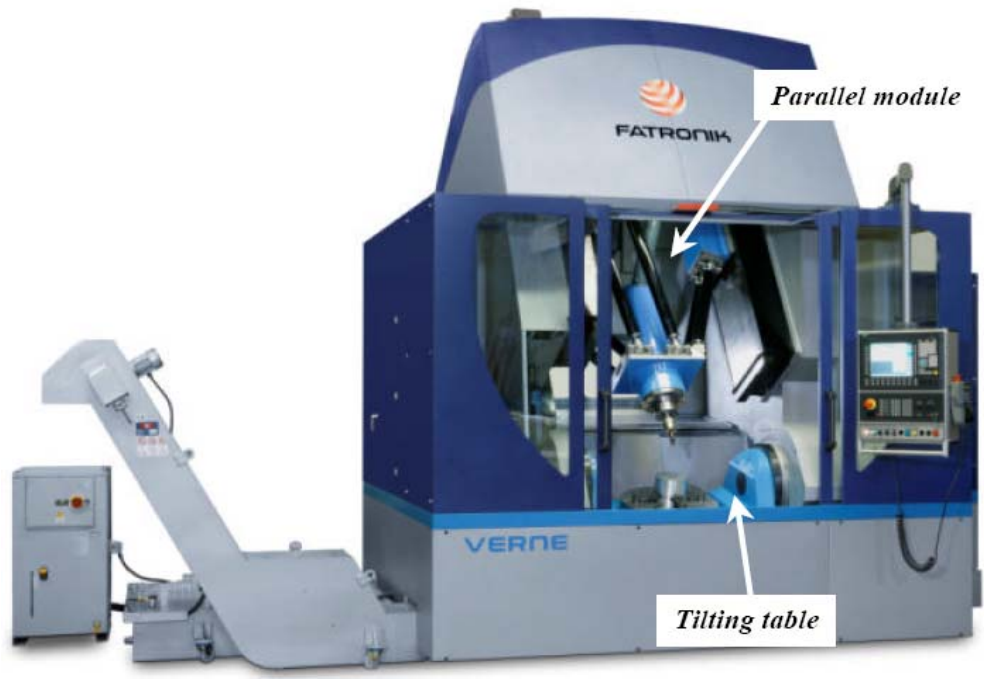

**Figure 1: Overall view of the VERNE machine**

The following section describes the VERNE machine. In section 3, we study the kinematics of the parallel module of the VERNE machine. In section 4 the methods presented in section 3 are extended to study the kinematic of the full VERNE machine. Finally Section 5 concludes this paper.

## 2. Description of the VERNE machine

The VERNE machine consists of a parallel module and a tilting table as shown in Fig. 2. The vertices of the moving platform of the parallel module are connected to a fixed-base plate through three legs I, II and III. Each leg uses a pair of rods linking a prismatic joint to the moving platform through two pairs of spherical joints. Legs II and III are two identical parallelograms. Leg I differs from the other two legs in that it is a trapezium instead of a parallelogram, namely, $A_{11}A_{12} \neq B_{11}B_{12}$, where $A_{ij}$ (respectively $B_{ij}$) is the center of spherical joint number j on the prismatic joint number i (respectively on the moving platform side), i = 1..3, j = 1..2. The movement of the moving platform is generated by three sliding actuators along three vertical guideways.

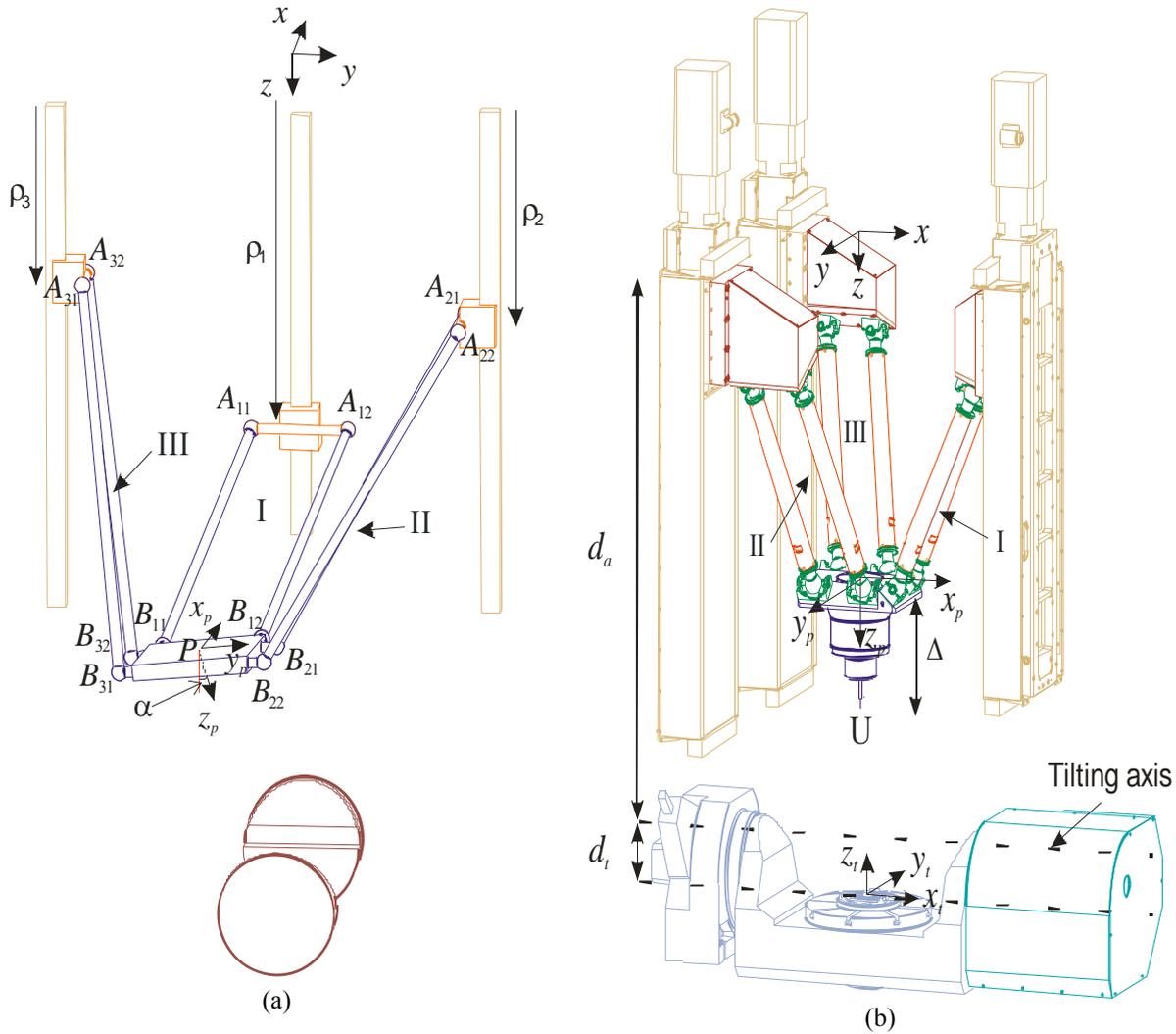

**Figure 2: Schematic representation of the VERNE machine; (a) simplified representation and (b) the real representation supplied by Fatronik**

Due to the arrangement of the links and joints, legs II and III prevent the platform from rotating about y and z axes. Leg I prevents the platform from rotating about z-axis (Fig. 2). Because this leg is a trapezium ( $A_{11}A_{12} \neq B_{11}B_{12}$ ), however, a slight coupled rotation $\alpha$ about the x-axis exists as shown in Fig. 2a. As shown further on, this coupled rotation makes the kinematic analysis more complex. Its impact on the workspace has not been fully investigated yet. The reasons why Fatronik has equipped leg I with a trapezium rather than with a parallelogram like in conventional linear Delta machines are beyond the authors' knowledge.

The tilting table is used to rotate the workpiece about two orthogonal axes. The first one, the tilting axis, is horizontal and the second one, the rotary axis, is always perpendicular to the tilting table.

This machine takes full advantage of these two additional axes to adjust the tool orientation with respect to the workpiece.

## 3. Kinematic analysis of the parallel module of the VERNE machine

### 3.1 Kinematic equations

In order to analyze the kinematics of our parallel module, two relative coordinates are assigned as shown in Fig. 2a. A static Cartesian frame $R_b = (O, x, y, z)$ is fixed at the base of the machine tool, with the z-axis pointing downward along the vertical direction. The mobile Cartesian frame, $R_{pl} = (P, x_P, y_P, z_P)$, is attached to the moving platform at point P.

In any constrained mechanical system, joints connecting bodies restrict their relative motion and impose constraints on the generalized coordinates, geometric constraints are then formulated as algebraic expressions involving generalized coordinates.

Let us $^b T_{pl}$ define the transformation matrix that brings the fixed Cartesian frame $R_b$ on the frame $R_{pl}$ linked to the moving platform.

$$^b T_{pl} = Trans(x_p, y_p, z_p) Rot(x, \alpha) \tag{1}$$

We use this transformation matrix to express $B_{ij}$ as function of $x_p$, $y_p$, $z_p$ and $\alpha$ by using the relation $B_{ij} = {}^b T_{pl} {}^{pl} B_{ij}$ where $^{pl} B_{ij}$ represents the point $B_{ij}$ expressed in the frame $R_{pl}$.

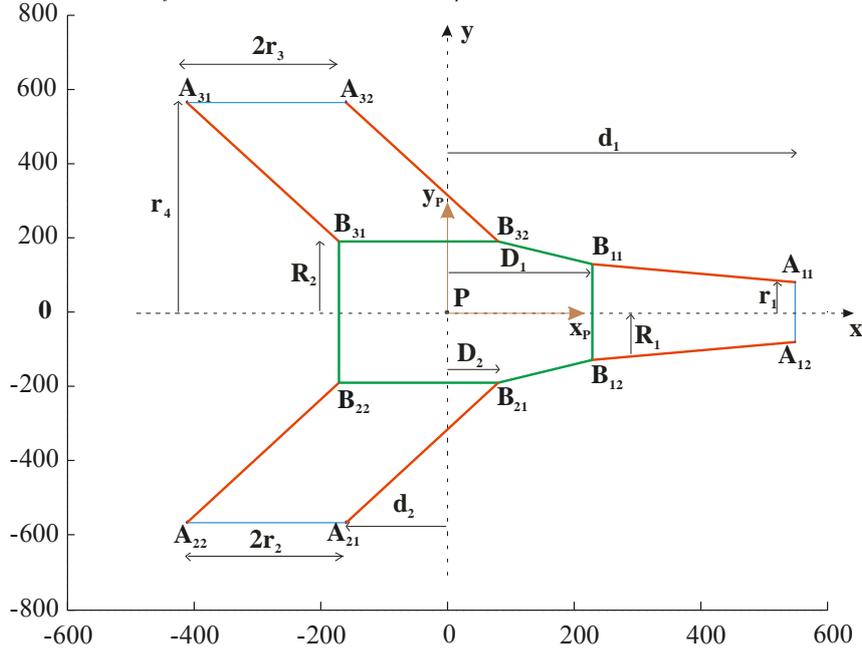

Figure 3: Dimensions of the parallel kinematic structure in the frame supplied by Fatronik

Using the parameters defined in Figs. 2 and 3, the constraint equations of the parallel manipulator are expressed as:

$$\|A_{ij} B_{ij}\|^2 - L_i^2 = (x_{Bij} - x_{Aij})^2 + (y_{Bij} - y_{Aij})^2 + (z_{Bij} - z_{Aij})^2 - L_i^2 = 0 \quad (i = 1..3, j = 1..2) \tag{2}$$

Leg I is represented by two different Eqs. (3a-3b). This is due to the fact that $A_{11}A_{12} \neq B_{11}B_{12}$ (figure 3).

$$(x_P + D_1 - d_1)^2 + (y_P + R_1 \cos(\alpha) - r_1)^2 + (z_P + R_1 \sin(\alpha) - \rho_1)^2 - L_1^2 = 0 \tag{3a}$$

$$(x_P + D_1 - d_1)^2 + (y_P - R_1 \cos(\alpha) + r_1)^2 + (z_P - R_1 \sin(\alpha) - \rho_1)^2 - L_1^2 = 0 \tag{3b}$$

Leg II is represented by a single Eq. (4).

$$(x_P + D_2 - d_2)^2 + (y_P - R_2 \cos(\alpha) + r_4)^2 + (z_P - R_2 \sin(\alpha) - \rho_2)^2 - L_2^2 = 0 \tag{4}$$

Leg III, which is similar to leg II (figure 3), is also represented by a single Eq. (5).

$$(x_P + D_2 - d_2)^2 + (y_P + R_2 \cos(\alpha) - r_4)^2 + (z_P + R_2 \sin(\alpha) - \rho_3)^2 - L_3^2 = 0 \tag{5}$$

### 3.2 Coupling between the position and the orientation of the platform

The parallel module of the VERNE machine possesses three actuators and three degrees of freedom. However, there is a coupling between the position and the orientation angle of the platform. The object of this section is to study the coupling constraint imposed by leg I.

By eliminating $\rho_1$ from Eqs. (3a) and (3b), we obtain a relation (6) between $x_P$, $y_P$ and $\alpha$ independently of $z_P$.

$$R_1^2 \sin^2(\alpha)(x_P + D_1 - d_1)^2 + (r_1^2 - 2R_1 r_1 \cos(\alpha) + R_1^2) y_P^2 - R_1^2 \sin^2(\alpha)(L_1^2 - (R_1^2 + r_1^2 - 2R_1 r_1 \cos(\alpha))) = 0 \quad (6)$$

We notice that for a given $\alpha$, Eq. (6) represents an ellipse (7). The size of this ellipse is determined by $a$ and $b$, where $a$ is the length of the semi major axis and $b$ is the length of the semi minor axis.

$$\frac{(x_P + D_1 - d_1)^2}{a^2} + \frac{y_P^2}{b^2} = 1 \quad (7)$$

where 
$$\begin{cases} a = \sqrt{(L_1^2 - (R_1^2 + r_1^2 - 2R_1 r_1 \cos(\alpha)))} \\ b = \sqrt{\dfrac{R_1^2 \sin^2(\alpha)(L_1^2 - (R_1^2 + r_1^2 - 2R_1 r_1 \cos(\alpha)))}{(r_1^2 - 2R_1 r_1 \cos(\alpha) + R_1^2)}} \end{cases}$$

These ellipses define the locus of points reachable with the same orientation $\alpha$.

## 3.3 The Inverse kinematics

The inverse kinematics deals with the determination of the joint coordinates as function of the moving platform position. For the inverse kinematic problem of our spatial parallel manipulator, the position coordinates ($x_P$, $y_P$, $z_P$) are given but the coordinates $\rho_i$ ($i = 1..3$) of the actuated prismatic joints and the orientation angle $\alpha$ of the moving platform are unknown.

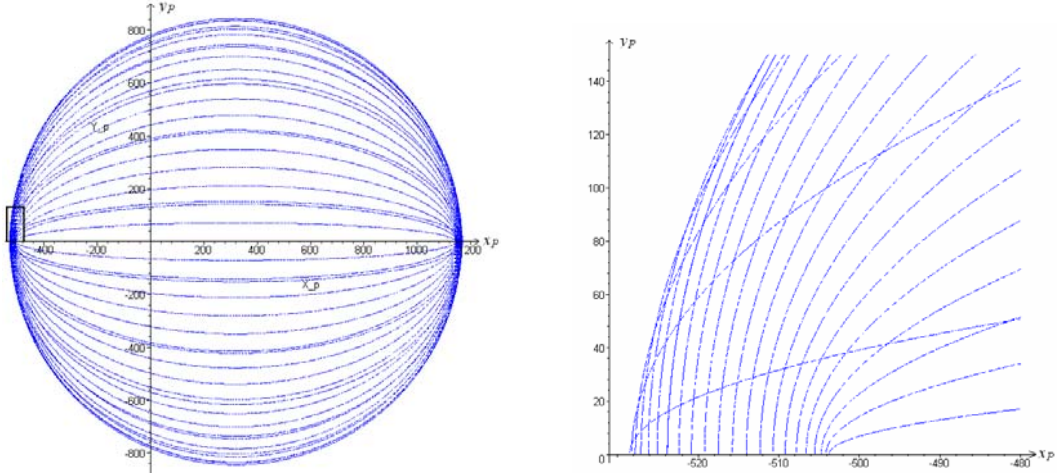

**Figure 4: (a) Curves of iso-values of the orientation $\alpha$ from $-\pi$ to $+\pi$ following a constant step of $2\pi/45$ (b) zoom of the framed zone**

To solve the inverse kinematic problem, we first find all the possible orientation angles $\alpha$ for prescribed values of the position of the platform ($x_P$, $y_P$, $z_P$). These orientations are determined by solving Eq. (8), a third-degree-characteristic polynomial in $\cos(\alpha)$ derived from Eq. (6).

$$p_1 \cos^3(\alpha) + p_2 \cos^2(\alpha) + p_3 \cos(\alpha) + p_4 = 0 \quad (8)$$

where 
$$\begin{cases} p_1 = 2R_1^3 r_1 \\ p_2 = R_1^2 (L_1^2 - R_1^2 - r_1^2) - R_1^2 (x_P + D_1 - d_1)^2 \\ p_3 = -2R_1^3 r_1 - 2R_1 r_1 y_P^2 \\ p_4 = R_1^2 (x_P + D_1 - d_1)^2 + (R_1^2 + r_1^2) y_P^2 - R_1^2 (L_1^2 - R_1^2 - r_1^2) \end{cases}$$

As shown in subsection 3.2, this equation also represents ellipses of iso-values of $\alpha$. So if we plot all ellipses together by varying $\alpha$ from $-\pi$ to $+\pi$ (figure 4), we notice that every point (defined by $x_P$, $y_P$ and $z_P$) is obtained by the intersection of two ellipses. Thus, each ellipse represents two opposite orientations so each point can have a maximum of four different orientations. This conclusion is verified by the fact that we can only find four real solutions to the polynomial (Table I).

| | |
|---|---|
| $\begin{cases} x_P, y_P, z_P \\ y_P \neq 0 \end{cases}$ | $\alpha = \{\pm\alpha_1 \text{ and } \pm\alpha_2\}$ |
| $\begin{cases} x_P, y_P, z_P \\ y_P = 0 \end{cases}$ | $\alpha = \{0, \pm\alpha_1, \pi\}$ |

**TABLE I: the possible orientations for a fixed position of the platform**

After finding all the possible orientations, we use the equations derived in subsection 3.1 to calculate the joint coordinates $\rho_i$ for each orientation angle $\alpha$. To make this task easier, we introduce two new points $A_1$ and $B_1$ as the middle of $A_{11}A_{12}$ and $B_{11}B_{12}$, respectively. The constraint equation of these two points is:

$$(x_P + D_1 - d_1)^2 + y_P^2 + (z_P - \rho_1)^2 - (L_1^2 - (R_1^2 + r_1^2 - 2R_1 r_1 \cos(\alpha))) = 0 \qquad (9)$$

Then, for prescribed values of the position and orientation of the platform, the required actuator inputs can be directly computed from equations (9), (4) and (5):

$$\rho_1 = z_P + s_1\sqrt{\left(\left(L_1^2 - (R_1^2 + r_1^2 - 2R_1 r_1 \cos(\alpha))\right) - (x_P + D_1 - d_1)^2 - y_P^2\right)} \qquad (10)$$

$$\rho_2 = z_P - R_2 \sin(\alpha) + s_2\sqrt{\left(L_2^2 - (x_P + D_2 - d_2)^2 - (y_P - R_2 \cos(\alpha) + r_4)^2\right)} \qquad (11)$$

$$\rho_3 = z_P + R_2 \sin(\alpha) + s_3\sqrt{\left(L_3^2 - (x_P + D_2 - d_2)^2 - (y_P + R_2 \cos(\alpha) - r_4)^2\right)} \qquad (12)$$

where $s_1, s_2, s_3 \in \{\pm 1\}$ are the configuration indices defined as the signs of $\rho_1 - z_P$, $\rho_2 - z_P + R_2 \sin(\alpha)$, $\rho_3 - z_P - R_2 \sin(\alpha)$, respectively.

Subtracting equation (3a) from equation (3b), yields:

$$y_P (R_1 \cos(\alpha) - r_1) = R_1 \sin(\alpha)(\rho_1 - z_P) \qquad (13)$$

Eq. (13) implies that: $\text{sgn}(\rho_1 - z_P)\text{sgn}(\sin(\alpha)) = \text{sgn}(R_1 \cos(\alpha) - r_1)\text{sgn}(y_P)$

This means that for prescribed values of the position and orientation of the platform, the joint coordinate $\rho_1$ possesses one solution, except when $\alpha = \{0, \pi\}$. In this case $s_1$ can take on both values $+1$ and $-1$. As a result $\rho_1$ can take on two values when $\alpha = \{0, \pi\}$.

| | |
|---|---|
| $\alpha = \{0, \pi\}$ | $s_1 = \pm 1$ |
| $\begin{cases} R_1 \cos(\alpha) = r_1 \\ y_P = 0 \text{ with } \alpha \neq 0 \end{cases}$ | $\rho_1 = z_P$ |
| others | $s_1 = +1$ or $-1$ |

**TABLE II. Solutions of the joint coordinate $\rho_1$ according to the values of $\alpha$**

Observing equations (10), (11), (12), Table I and Table II, we conclude that there are four solutions for leg I and two solutions for leg II and III. Thus there are sixteen inverse kinematic solutions for the parallel module (figure 5).

From the sixteen theoretical inverse kinematics solutions shown in figure 5, only one is used by the VERNE machine: the one referred to as (m) in figure 5, which is characterized by the fact that each leg must have its slider attachment points above the moving platform attachment points, i.e. $s_i = -1$ (remember that the z-axis is directed downward).

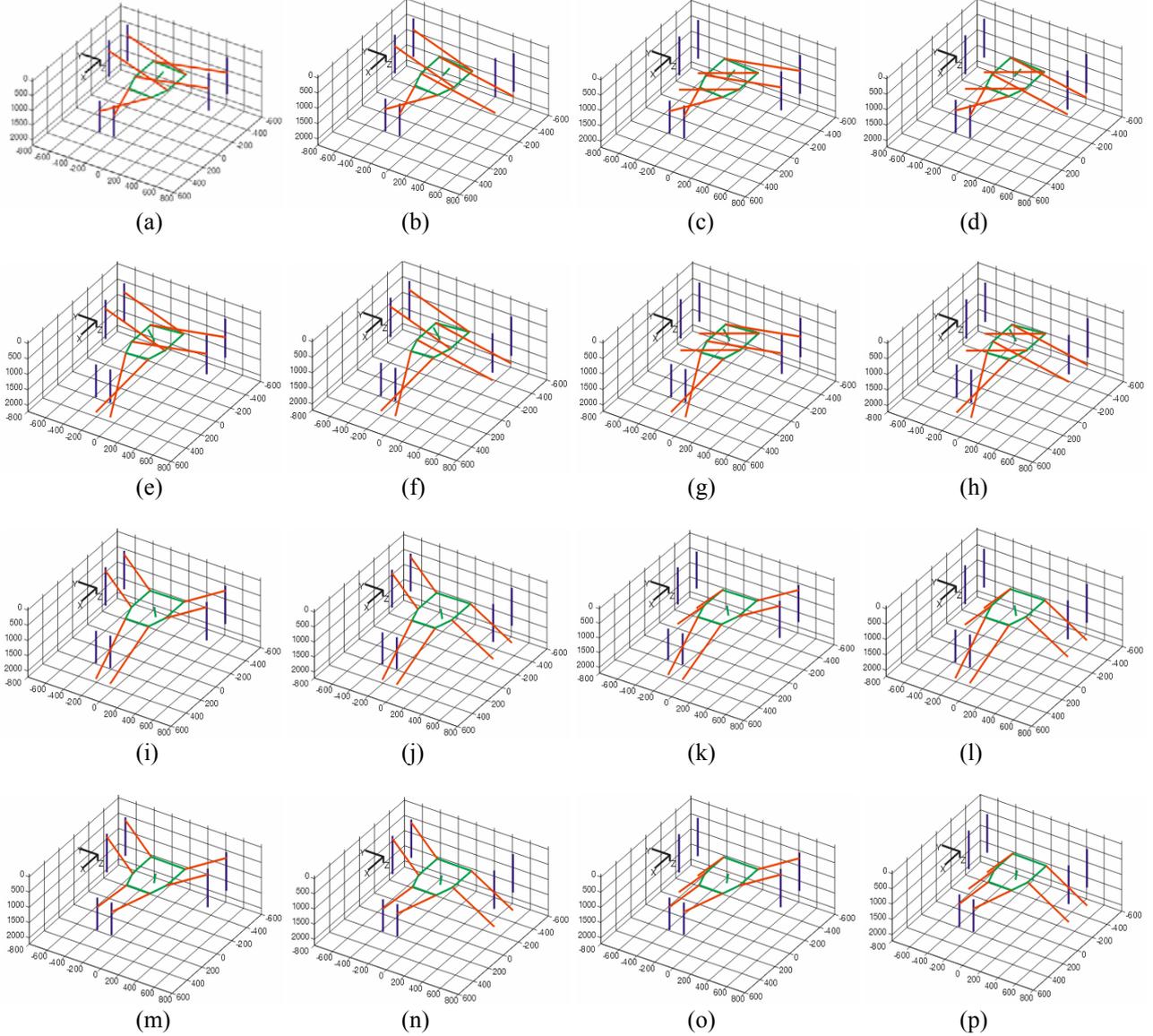

**Figure 5: The sixteen solutions to the inverse kinematics problem when $x_P$ = -240 mm, $y_P$ = -86 mm and $z_P$ = 1000 mm**

For the remaining 15 solutions one of the sliders leaves its joint limits or the two rods of leg I cross. Most of these solutions are characterized by the fact that at least one of the legs has its slider attachment points below the moving platform attachment points. So only $s_1, s_2, s_3 = -1$ in Eqs. (10-12) must be selected (remember that the z-axis is directed downward). To prevent rod crossing, we also add a condition on the orientation of the moving platform. This condition is $R_1 \cos(\alpha) > r_1$. Finally, we check the joint limits of the sliders as well as the serial singularities [15], [20].

For the VERNE parallel module, applying the above conditions will always yield a unique solution for practical applications (solution (m) shown in Fig. 5).

### 3.4 The forward kinematics

The forward kinematics deals with the determination of the moving platform position as function of the joint coordinates. For the forward kinematics of our spatial parallel manipulator, the values of the joint coordinates $\rho_i$ $(i = 1..3)$ are known and the goal is to find the coordinates $x_P$, $y_P$ and $z_P$ of the centre of the moving platform P.

To solve the forward kinematics, we eliminate successively $x_P$, $y_P$ and $z_P$ from the system $(S1)$ of four equations ((3a), (3b), (4) and (5)) to have an equation function of the joint coordinates $\rho_i$ ($i = 1..3$) and function of the orientation angle $\alpha$ of the platform. To do so, we first compute $y_P$ as function of $z_P$ in Eq. (14) by subtracting Eq. (3a) from Eq. (3b)

$$y_p = \frac{R_1 \sin(\alpha)(\rho_1 - z_p)}{(R_1 \cos(\alpha) - r_1)} \tag{14}$$

The expression of $y_p$ in Eq. (14) is substituted into system $(S1)$ to obtain a new system $(S2)$ of three Eqs. (15), (16) and (17) derived from Eqs. (3a), (4) and (5) respectively.

$$\begin{aligned}
& 2R_1^3 r_1 \cos^3(\alpha) + R_1^2\left((x_p + D_1 - d_1)^2 + R_1^2 + 5r_1^2 - L_1^2\right)\cos^2(\alpha) + (R_1^2 + r_1^2)(\rho_1 - z_p)^2 - \\
& 2R_1 r_1\left((x_p + D_1 - d_1)^2 + (\rho_1 - z_p)^2 + R_1^2 + 2r_1^2 - L_1^2\right)\cos(\alpha) + r_1^2\left((x_p + D_1 - d_1)^2 + R_1^2 + r_1^2 - L_1^2\right) = 0
\end{aligned} \tag{15}$$

$$\begin{aligned}
& 2R_1\left(R_1 r_4(\rho_1 - z_p) + R_2 r_1(\rho_1 - 2\rho_2 + z_p)\right)\cos(\alpha)\sin(\alpha) + 2r_1\left(R_2 r_1(\rho_2 - z_p) - R_1 r_4(\rho_1 - z_p)\right)\sin(\alpha) + \\
& R_1\left(R_1\left((x_p + D_2 - d_1)^2 - (\rho_1 - z_p)^2 + (\rho_2 - z_p)^2 + R_2^2 + r_4^2 - L_2^2\right) + 4R_2 r_4 r_1\right)\cos^2(\alpha) - \\
& \left(2R_2 r_1^2 r_4 + 2R_1 r_1\left((x_p + D_2 - d_2)^2 + (\rho_2 - z_p)^2 + R_2^2 + r_4^2 - L_2^2\right)\right)\cos(\alpha) + 2R_1^2 R_2(\rho_2 - \rho_1)\sin(\alpha)\cos^2(\alpha) + \\
& R_1^2(\rho_1 - z_p)^2 + r_1^2\left((x_p + D_2 - d_2)^2 + (\rho_2 - z_p)^2 + R_2^2 + r_4^2 - L_2^2\right) - 2R_1^2 R_2 r_4 \cos^3(\alpha) = 0
\end{aligned} \tag{16}$$

$$\begin{aligned}
& 2R_1\left(-R_1 r_4(\rho_1 - z_p) - R_2 r_1(\rho_1 - 2\rho_3 + z_p)\right)\cos(\alpha)\sin(\alpha) + 2r_1\left(-R_2 r_1(\rho_3 - z_p) + R_1 r_4(\rho_1 - z_p)\right)\sin(\alpha) + \\
& R_1\left(R_1\left((x_p + D_2 - d_1)^2 - (\rho_1 - z_p)^2 + (\rho_3 - z_p)^2 + R_2^2 + r_4^2 - L_3^2\right) + 4R_2 r_4 r_1\right)\cos^2(\alpha) - \\
& \left(2R_2 r_1^2 r_4 + 2R_1 r_1\left((x_p + D_2 - d_2)^2 + (\rho_3 - z_p)^2 + R_2^2 + r_4^2 - L_3^2\right)\right)\cos(\alpha) - 2R_1^2 R_2(\rho_3 - \rho_1)\sin(\alpha)\cos^2(\alpha) + \\
& R_1^2(\rho_1 - z_p)^2 + r_1^2\left((x_p + D_2 - d_2)^2 + (\rho_3 - z_p)^2 + R_2^2 + r_4^2 - L_3^2\right) - 2R_1^2 R_2 r_4 \cos^3(\alpha) = 0
\end{aligned} \tag{17}$$

We then compute $z_P$ as function of $\rho_i$ ($i = 1..3$) and $\alpha$ in Eq. (18) by subtracting equation (16) from equation (17).

$$z_p = \frac{\left((R_1 \cos(\alpha) - r_1)((\rho_2 + \rho_3)(\rho_3 - \rho_2) - 2R_2(\rho_3 + \rho_2 - 2\rho_1)\sin(\alpha)) + 4C_1 \rho_1 \sin(\alpha)\right)}{2(2C_1 \sin(\alpha) + (R_1 \cos(\alpha) - r_1)(\rho_3 - \rho_2))} \tag{18}$$

where $C_1 = (r_1 R_2 - r_4 R_1)$

The expression of $z_p$ in Eq. (18) is substituted into system $(S2)$ to obtain a new system $(S3)$ of two equations (19) and (20) derived from equations (15) and (16) respectively. Finally, we compute $x_p$ as function of $\rho_i$ ($i = 1..3$) and $\alpha$ by subtracting equation (19) from equation (20).

$$x_p = \frac{\begin{aligned}& -2R_2 C_1(\rho_3 - \rho_2)\sin^2(\alpha) + \\ & \left(C_1\left(2C_2 + (\rho_3 - \rho_1)^2 + (\rho_2 - \rho_1)^2 - 4r_4 R_2 \cos(\alpha)\right) + 4(r_1 C_1 + R_2(\rho_3 - \rho_1)(\rho_2 - \rho_1))(R_1 \cos(\alpha) - r_1)\right)\sin(\alpha) + \\ & (\rho_3 - \rho_2)\left(C_2 - (\rho_3 - \rho_1)(\rho_2 - \rho_1) - 2r_1^2 + 2(r_1 R_1 - r_4 R_2)\cos(\alpha)\right)(R_1 \cos(\alpha) - r_1)\end{aligned}}{2(D_1 - d_1 - D_2 + d_2)\left(2C_1 \sin(\alpha) + (\rho_3 - \rho_2)(R_1 \cos(\alpha) - r_1)\right)} \tag{21}$$

where $C_2 = (D_2 - d_2)^2 - (D_1 - d_1)^2 + r_1^2 + r_4^2 - R_1^2 + R_2^2 + L_1^2 - L_3^2$

Then the above expression of $x_p$ is substituted into system $(S3)$.

The resulting equations of system $(S3)$ are given in Appendix A.

For each step, we determine solution existence conditions by studying the denominators that appear in the expressions of $x_P$, $y_P$ and $z_P$. These conditions are:

$$R_1 \cos(\alpha) - r_1 \neq 0 \quad (22)$$

$$2C_1 \sin(\alpha) + (\rho_3 - \rho_2)(R_1 \cos(\alpha) - r_1) \neq 0 \quad (23)$$

Equation (22) obtained from (13) implies that $A_1B_1$ is perpendicular to the slider plane of leg I. In this case equation (7) represents a circle because $a = b$.

When $\rho_2 = \rho_3$ in equation (23), we have $\alpha = \{0, \pi\}$. This means that $y_P = 0$ (obtained from Equations. (4) − (5)).

To finish the resolution of the system, we perform the tangent-half-angle substitution $t = \tan(\alpha/2)$. As a consequence, the forward kinematics of our parallel manipulator results in a eight-degree-characteristic polynomial in $t$, whose coefficients are relatively large expressions in $\rho_1$, $\rho_2$ and $\rho_3$. Expressions of these coefficients are not reported here because of space limitation. They are available in [20]. Knowing the value of $\alpha$, we calculate $x_p$, $y_p$ and $z_p$ using Eqs (21), (14) and (18), respectively. For the VERNE machine, only 4 assembly-modes have been found (figure 6). It was possible to find up to 6 assembly-modes but only for input joint values out of the reachable joint space of the machine.

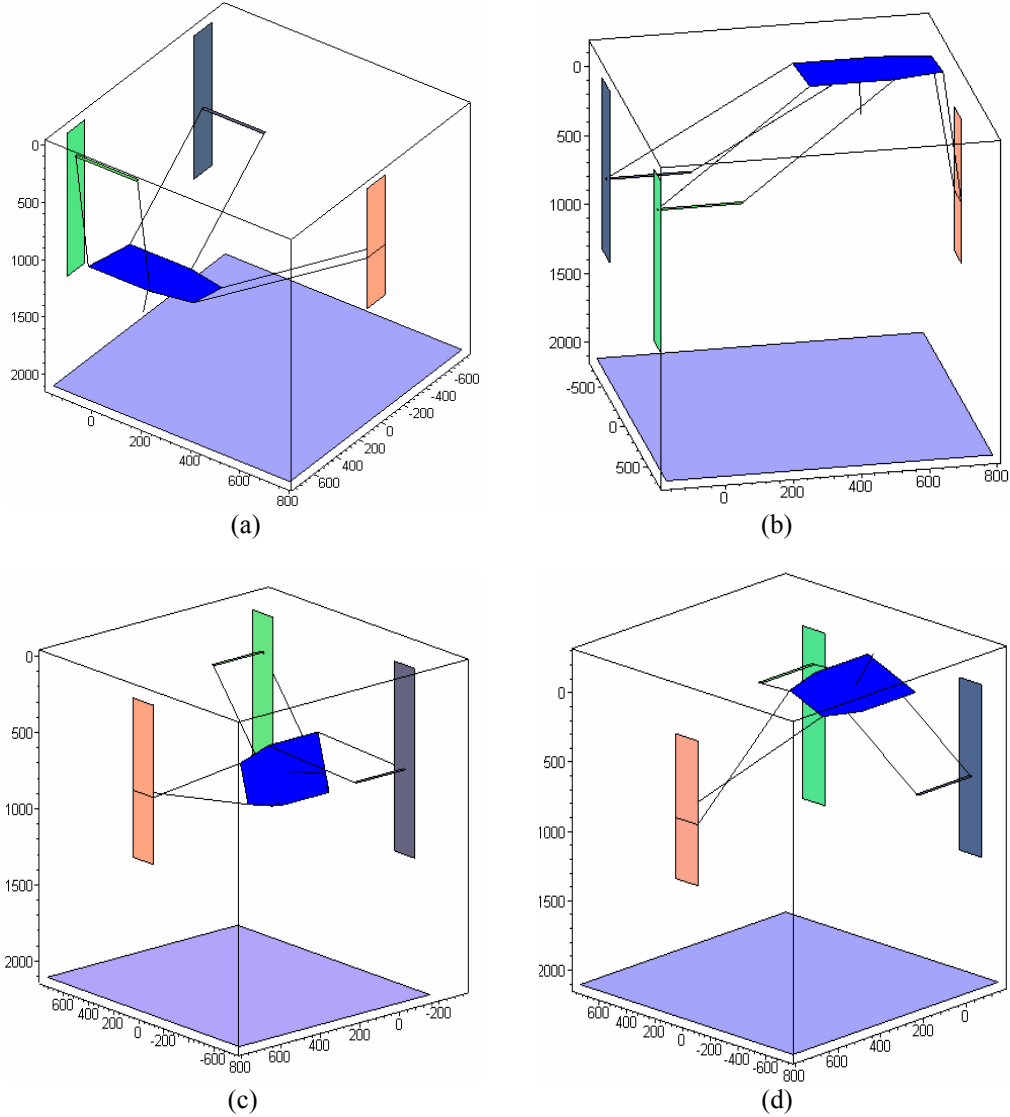

(a)  (b)

(c)  (d)

**Figure 6: The four assembly-modes of the VERNE parallel module for $\rho_1 = 674$ mm, $\rho_2 = 685$ mm and $\rho_3 = 250$ mm. only (a) is reachable by the actual machine**

Only one assembly-mode is actually reachable by the machine (solution (a) shown in Fig. 6) because the other ones lead to either rod crossing, collisions, or joint limit violation. The right assembly mode can be recognized, like for the right working mode, by the fact that each leg must have its slider attachment points above the moving platform attachment points, i.e. $s_i = -1$ (keep in mind that the z-axis is directed downwards).

The proposed method for calculating the various solutions of the forward kinematic problem has been implemented in Maple. Table III give the solutions for $\rho_1 = 674$ mm, $\rho_2 = 685$ mm, $\rho_3 = 250$ mm and Fig. 6 shows the four assembly modes

| $\rho_1 = 674$ mm, $\rho_2 = 685$ mm and $\rho_3 = 250$ mm | | | | |
|---|---|---|---|---|
| Case | $\alpha$ (rd) | $x_P$ (mm) | $y_P$ (mm) | $z_P$ (mm) |
| (a) | -0.22 | -199.80 | 355.92 | 1242 |
| (b) | -0.14 | 298.35 | -297.53 | -120.22 |
| (c) | 1.81 | -393.6 | 322.82 | 958.21 |
| (d) | 2.70 | -115.62 | -189.68 | -0.26 |

**TABLE III: the numerical results of the forward kinematic problem of the example where $\rho_1 = 674$ mm, $\rho_2 = 685$ mm and $\rho_3 = 250$ mm**

## 4. Kinematic analysis of the full VERNE machine (parallel module + tilting table)

### 4.1 Kinematic equations

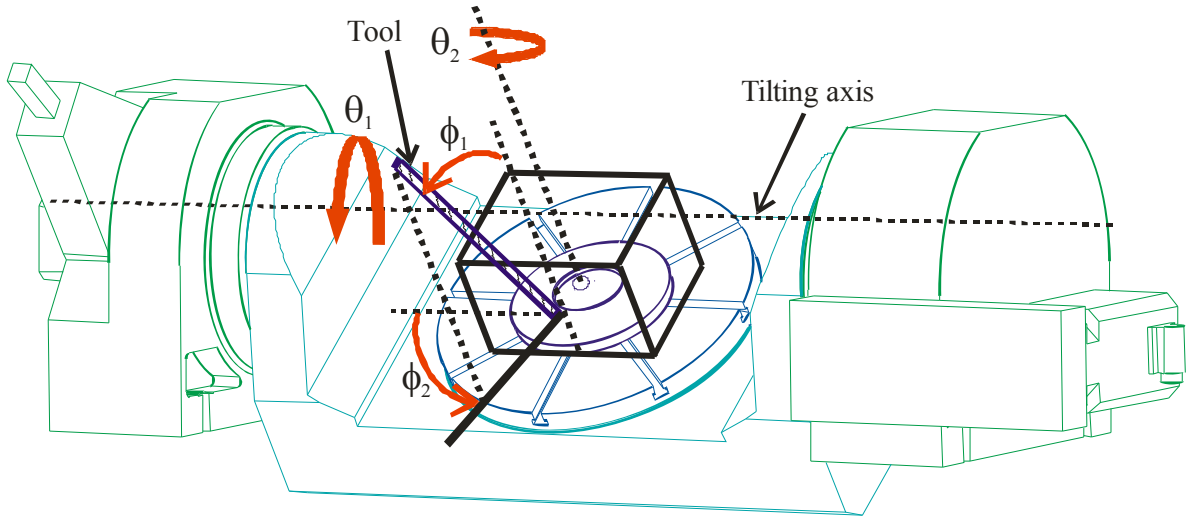

**Figure 7: Draw of the tilting table: the tool orientation is defined by two angles ($\phi_1$, $\phi_2$) relative to frame $R_t$ linked to the tilting table. The orientation angles ($\theta_1, \theta_2$) of the tilting table are defined relative to frame $R_b$ fixed to the base of the VERNE machine**

In order to analyze the kinematics of the VERNE machine, we define the following coordinate frame as shown below in Table IV:

| Transformation | Axis | Angles/Distance | Input Frame | Output Frame |
|---|---|---|---|---|
| Translation | z | $d_a$ | $R_b(O, x, y, z)$ | $R_1(O_1, x_1, y_1, z_1)$ |
| Rotation | $x_1$ | $\theta_1$ | $R_1(O_1, x_1, y_1, z_1)$ | $R_2(O_2, x_2, y_2, z_2)$ |
| Translation | $z_2$ | $d_t$ | $R_2(O_2, x_2, y_2, z_2)$ | $R_3(O_3, x_3, y_3, z_3)$ |
| Rotation | $x_3$ | $\pi$ | $R_3(O_3, x_3, y_3, z_3)$ | $R_4(O_4, x_4, y_4, z_4)$ |
| Rotation | $z_4$ | $\theta_2$ | $R_4(O_4, x_4, y_4, z_4)$ | $R_t(t, x_t, y_t, z_t)$ |
| Translation | $x_t, y_t, z_t$ | $x_u, y_u, z_u$ | $R_t(t, x_t, y_t, z_t)$ | $R_5(O_5, x_5, y_5, z_5)$ |
| Rotation | $z_5$ | $\phi_2$ | $R_5(O_5, x_5, y_5, z_5)$ | $R_6(O_6, x_6, y_6, z_6)$ |
| Rotation | $x_6$ | $\pi + \phi_1$ | $R_6(O_6, x_6, y_6, z_6)$ | $R_7(O_7, x_7, y_7, z_7)$ |
| Translation | $z_7$ | $-\Delta$ | $R_7(O_7, x_7, y_7, z_7)$ | $R_{pl}(P, x_p, y_p, z_p)$ |

**Table IV:** Transformation matrices that bring the input frame on the output frame; where $x_u$, $y_u$ and $z_u$ are the coordinates of the tool centre point (TCP), U, in $R_t$

Let ${}^bT_t$ define the transformation matrix that brings the fixed Cartesian frame $R_b$ on the frame $R_t$ linked to the tilting table.

$$^bT_t = trans(z, d_a) rot(x_1, \theta_1) trans(z_2, d_t) rot(x_3, \pi) rot(z_4, \theta_2) \qquad (24)$$

Let ${}^tT_{pl}$ define the transformation matrix that brings the frame $R_t$ linked to the tilting table on the frame $R_{pl}$ linked to the moving platform.

$$^tT_{pl} = trans(x_u, y_u, z_u) rot(z_5, \phi_2) rot(x_6, \pi + \phi_1) trans(z_7, -\Delta) \qquad (25)$$

We use transformation matrices from Eqs. (24) and (25) in order to express $B_{ij}$ as function of $x_u$, $y_u$, $z_u$, $\phi_1$, $\phi_2$, $\theta_1$ and $\theta_2$ by using the relation $B_{ij} = {}^bT_{pl}{}^{pl}B_{ij}$ where ${}^bT_{pl} = {}^bT_t{}^tT_{pl}$ and ${}^{pl}B_{ij}$ represent the point $B_{ij}$ expressed in the frame $R_{pl}$. Using Eq. (2) from section 3.1 and the parameters defined in Figs. 2 and 3, we can express all constraint equations of the VERNE machine. However knowing that $A_{i1}B_{i1}$ and $A_{i2}B_{i2}$ are parallel for i=1..2, we can prove that

$$\theta_2 = -\phi_2 \qquad (26)$$

Substituting the above value of $\theta_2$ in all constraint equations resulting from Eq. (2), we obtain that leg I is represented by two different equations (27a) and (27b) while leg II (respectively leg III) is represented by only one equation (28) (respectively equation (29)).

$$\begin{aligned}&\left(\cos(\phi_2) x_u + \sin(\phi_2) y_u + D_1 - d_1\right)^2 + \\ &\left(\sin(\theta_1)(z_u - d_t) + \cos(\theta_1)(\sin(\phi_2) x_u - \cos(\phi_2) y_u) + \Delta\sin(\theta_1 + \phi_1) + R_1 \cos(\theta_1 + \phi_1) - r_1\right)^2 + \\ &\left(\sin(\theta_1)(\sin(\phi_2) x_u - \cos(\phi_2) y_u) - \cos(\theta_1)(z_u - d_t) + d_a - \Delta\cos(\theta_1 + \phi_1) + R_1 \sin(\theta_1 + \phi_1) - \rho_1\right)^2 - L_1^2 = 0\end{aligned} \qquad (27a)$$

$$\begin{aligned}&\left(\cos(\phi_2) x_u + \sin(\phi_2) y_u + D_1 - d_1\right)^2 + \\ &\left(\sin(\theta_1)(z_u - d_t) + \cos(\theta_1)(\sin(\phi_2) x_u - \cos(\phi_2) y_u) + \Delta\sin(\theta_1 + \phi_1) - R_1 \cos(\theta_1 + \phi_1) + r_1\right)^2 + \\ &\left(\sin(\theta_1)(\sin(\phi_2) x_u - \cos(\phi_2) y_u) - \cos(\theta_1)(z_u - d_t) + d_a - \Delta\cos(\theta_1 + \phi_1) - R_1 \sin(\theta_1 + \phi_1) - \rho_1\right)^2 - L_1^2 = 0\end{aligned} \qquad (27b)$$

$$\begin{aligned}&\left(\cos(\phi_2) x_u + \sin(\phi_2) y_u + D_2 - d_2\right)^2 + \\ &\left(\sin(\theta_1)(z_u - d_t) + \cos(\theta_1)(\sin(\phi_2) x_u - \cos(\phi_2) y_u) + \Delta\sin(\theta_1 + \phi_1) - R_2 \cos(\theta_1 + \phi_1) + r_4\right)^2 + \\ &\left(\sin(\theta_1)(\sin(\phi_2) x_u - \cos(\phi_2) y_u) - \cos(\theta_1)(z_u - d_t) + d_a - \Delta\cos(\theta_1 + \phi_1) - R_2 \sin(\theta_1 + \phi_1) - \rho_2\right)^2 - L_2^2 = 0\end{aligned} \qquad (28)$$

$$\begin{aligned}&\left(\cos(\phi_2)\, x_u + \sin(\phi_2)\, y_u + D_2 - d_2\right)^2 + \\ &\left(\sin(\theta_1)(z_u - d_t) + \cos(\theta_1)\left(\sin(\phi_2)\, x_u - \cos(\phi_2)\, y_u\right) + \Delta \sin(\theta_1 + \phi_1) + R_2 \cos(\theta_1 + \phi_1) - r_4\right)^2 + \\ &\left(\sin(\theta_1)\left(\sin(\phi_2)\, x_u - \cos(\phi_2)\, y_u\right) - \cos(\theta_1)(z_u - d_t) + d_a - \Delta \cos(\theta_1 + \phi_1) + R_2 \sin(\theta_1 + \phi_1) - \rho_3\right)^2 - L_3^2 = 0\end{aligned} \quad (29)$$

Identification of Eqs. (27a), (27b), (28) and (29) with Eqs. (3a), (3b), (4) and (5) respectively, yields:

$$\alpha = \theta_1 + \phi_1 \quad (30)$$

Condition (30) will help us understand the behavior of the VERNE machine from the one already studied in section 3 for its parallel module.

## 4.2 The inverse kinematics

For the inverse kinematic problem of the VERNE machine, the position of the TCP ($x_u$, $y_u$, $z_u$) and the orientation of the tool ($\phi_1$ and $\phi_2$) are given relative to frame $R_t$, but the joint coordinates, defined by the position $\rho_i$ ($i = 1..3$) of the actuated prismatic and the orientation ($\theta_1$ and $\theta_2$) of the tilting table in the base frame $R_b$ are unknown.

Knowing that $\theta_2 = -\phi_2$ from (26), the problem consists in solving the system ($S4$) of 4 equations ((27a), (27b), (28) and (29)) for only 4 unknowns ($\rho_i$ ($i = 1..3$) and $\theta_1$).

To solve the inverse kinematics, we follow the same reasoning as in subsection 3.3. First, we eliminate $\rho_1$ from Eqs. (27a) and (27b) in order to obtain a relation (31) between the TCP position and orientation ($x_u$, $y_u$, $z_u$, $\phi_1$ and $\phi_2$) and the tilting angle $\theta_1$.

$$\begin{aligned}&R_1^2 \sin^2(\theta_1 + \phi_1)\left(\cos(\phi_2)\, x_u + \sin(\phi_2)\, y_u + D_1 - d_1\right)^2 + \\ &\left(r_1^2 - 2R_1 r_1 \cos(\theta_1 + \phi_1) + R_1^2\right)\left(\sin(\theta_1)(z_u - d_t) + \cos(\theta_1)\left(\sin(\phi_2)\, x_u - \cos(\phi_2)\, y_u\right) + \Delta \sin(\theta_1 + \phi_1)\right)^2 - \\ &R_1^2 \sin^2(\theta_1 + \phi_1)\left(L_1^2 - \left(R_1^2 + r_1^2 - 2R_1 r_1 \cos(\theta_1 + \phi_1)\right)\right) = 0\end{aligned} \quad (31)$$

Then, we find all possible orientation angles $\theta_1$ for prescribed values of the position and the orientation of the tool. These orientations are determined by solving a six-degree-characteristic polynomial in $\tan(\theta_1/2)$ derived from Eq. (31). This polynomial can have up to four real solutions. This conclusion is verified by the fact that $\theta_1 = \phi_1 - \alpha$ from Eq. 30 where $\alpha$ can have only four real solutions as proved in subsection 3.3. After finding all the possible orientations, we use the system of equations ($S4$) in order to calculate the joint coordinates $\rho_i$ for each orientation angle $\theta_1$.

For $\rho_1$, we must verify that the values of $\rho_1$ obtained from Eqs. (27a) and (27b) are the same, as a result, we eliminate one of the two solutions.

Observing the above remark and equations (27a-27b), (28), (29) defined as two-degree-polynomials in $\rho_i$, $i = 1..3$ respectively, we conclude that there are four solutions for leg I and two solutions for leg II and III. Thus there are sixteen inverse kinematic solutions for the VERNE machine.

As above, from the sixteen theoretical inverse kinematics solutions, only one is used by the VERNE machine. This solution is characterized by the fact that each leg must have its slider attachment points above the moving platform attachment points.

For the remaining 15 solutions one of the sliders leaves its joint limits or the two rods of leg I cross. Most of these solutions are characterized by the fact that at least one of the legs has its slider attachment points lower than the moving platform attachment points. To prevent rod crossing, we also add a condition on the orientation of the moving platform. This condition is $R_1 \cos(\theta_1 + \phi_1) > r_1$. Finally, we check the joint limits of the sliders and the serial singularities [15].

As already mentioned, applying the above conditions will always yield to a unique solution for practical applications.

## 4.3 The forward kinematics

For the forward kinematics of the VERNE machine, the values of the joint coordinates, defined by the position $\rho_i$ ($i=1..3$) of the actuated prismatic and the orientation ($\theta_1$ and $\theta_2$) of the tilting table in the base frame $R_b$ are known and the goal is to find the position of the TCP ($x_u$, $y_u$, $z_u$) and the orientation of the tool ($\phi_1$ and $\phi_2$) in the frame $R_t$. Knowing that $\phi_2 = -\theta_2$ from (26) and $\phi_1 = \alpha - \theta_1$ from (30), we solve this problem by first solving the forward kinematics of the parallel module of the VERNE machine in order to find the coordinates $x_P$, $y_P$ and $z_P$ of the centre of the moving platform P and the orientation $\alpha$ of the moving platform in term of the joint coordinates $\rho_i$ ($i=1..3$). We then use transformation matrices from Eqs. (1) and (24) in order to express the tool position and orientation ($x_u$, $y_u$, $z_u$, $\phi_1$ and $\phi_2$) as function of $(x_P, y_P, z_P, \theta_1, \theta_2)$.

$$^tU = {}^tT_b^{\,b}U = {}^tT_b^{\,b}T_{pl}^{\,pl}U = {}^bT_t^{-1}T_{pl}^{\,pl}U \tag{32}$$

where $^{pl}U = \begin{bmatrix} 0 & 0 & \Delta & 1 \end{bmatrix}^T$ and $^tU = \begin{bmatrix} x_u & y_u & z_u & 1 \end{bmatrix}^T$ represent the TCP, $U$, expressed in frames $R_{pl}$ (linked to the moving platform) and the base frame $R_b$ respectively. Finally we obtain:

$$\begin{cases} \phi_1 = \alpha - \theta_1 \\ \phi_2 = -\theta_2 \\ x_u = \cos(\theta_2)x_p + \sin(\theta_2)\left(\Delta\sin(\alpha-\theta_1) - \cos(\theta_1)y_p - \sin(\theta_1)(z_p - d_a)\right) \\ y_u = -\sin(\theta_2)x_p + \cos(\theta_2)\left(\Delta\sin(\alpha-\theta_1) - \cos(\theta_1)y_p - \sin(\theta_1)(z_p - d_a)\right) \\ z_u = \sin(\theta_1)y_p - \cos(\theta_1)z_p + d_a\cos(\theta_1) - \Delta\cos(\alpha-\theta_1) + d_t \end{cases} \tag{33}$$

The VERNE machine behaves like its parallel module, so only 4 assembly-modes is found (figure 6) and only one assembly-mode is actually reachable by the machine (solution (a) shown in Fig. 6).

The proposed method for calculating the various solutions of the forward kinematic problem has been implemented in Maple. Table V give the solution for $\rho_1 = 674$ mm, $\rho_2 = 685$ mm, $\rho_3 = 250$ mm, $\theta_1 = 0.19$ rd and $\theta_2 = 0.39$ rd, the corresponding assembly modes for the parallel module were shown in Fig. 6.

| $\rho_1 = 674$ mm, $\rho_2 = 685$ mm, $\rho_3 = 250$ mm, $\theta_1 = 0.19$ rd and $\theta_2 = 0.39$ rd | | | | | |
|---|---|---|---|---|---|
| Case | $\phi_1$ rd | $\phi_2$ rd | $x_u$ (mm) | $y_u$ (mm) | $z_u$ (mm) |
| (a) | -0.41 | -0.39 | -338.06 | -296.89 | 461.6 |
| (b) | -0.33 | -0.39 | 478.52 | 379.38 | 1661.55 |
| (c) | 1.62 | -0.39 | -22106. | 497.49 | 1213.31 |
| (d) | 2.51 | -0.39 | 219.2 | 837.37 | 2433.67 |

**TABLE V: the numerical results of the forward kinematic problem of the example where $\rho_1 = 674$ mm, $\rho_2 = 685$ mm, $\rho_3 = 250$ mm, $\theta_1 = 0.19$ rd and $\theta_2 = 0.39$ rd**

## 5. Conclusion

This paper was devoted to the kinematic analysis of a 5-DOF hybrid machine tool, the VERNE machine. This machine possesses a complex motion caused by the unsymmetrical architecture of the parallel module where one of the legs is different from the other two legs. The inverse kinematics and the different assembly modes were derived. The forward kinematics was solved with the substitution method. It was shown that the inverse kinematics has sixteen solutions and the forward kinematics may have six real solutions. Examples were provided to illustrate the results. The special geometry of one of the legs highly complicates the kinematic models. Because two of the opposite sides of this leg have different lengths, the leg does not remain planar (rod directions define skew lines) as the machine moves, unlike what arises in the other two legs that are articulated parallelograms. As a result, a coupling angle of the moving platform about the x-axis exists. The derivation of the inverse and forward kinematic equations was not a trivial task and required much effort. This work is of interest as it may improve the control of the machine. It is worth noting that the VERNE machine is currently used every day for machining complex parts, especially for the molding industry. It is thus important to try to improve the efficiency of the machine. The controller of the actual VERNE machine resorts to an iterative Newton-Raphson resolution of the kinematic models. A fully comparative study between the symbolic and the iterative approach is still in progress and will be presented in forthcoming publications. It is expected that the symbolic method could decrease the Cpu-time and improve the quality of the control. The symbolic equations derived in this work are currently implemented in a simulation package of PKMs.

## 6. Appendix A

$$\begin{aligned}
& 8r_1R_1\left(\left(R_2^2 - R_1^2\right)(\rho_3 - \rho_2)^2 - 4R_2^2\left((\rho_2 - \rho_1)(\rho_3 - \rho_2) + (\rho_2 - \rho_1)^2\right) + 4(r_1R_2 - r_4R_1)^2\right)\cos^3(\alpha) + \\
& 32r_1R_1^2(\rho_3 - \rho_2)(r_4R_1 - r_1R_2)\cos^2(\alpha)\sin(\alpha) + \\
& \left(4(\rho_3 - \rho_2)^2\left((x_p + D_1 - d_1)^2 R_1^2 + 5R_1^2r_1^2 - R_2^2r_1^2 - R_1^2R_2^2 - R_1^2L_1^2 + R_1^4\right) - 16R_2^2(\rho_2 - \rho_1)(\rho_3 - \rho_2)(r_1^2 + R_1^2) - \right. \\
& 16(x_p + D_1 - d_1)^2(r_1R_2 - r_4R_1)^2 + 16r_4R_2R_1\left(2r_1^3 - 2r_1L_1^2 + 2r_1R_1^2\right) - 16R_1^2R_2^2\left((\rho_2 - \rho_1)^2 + r_1^2\right) - \\
& \left. 16r_1^2R_2^2\left((\rho_2 - \rho_1)^2 + r_1^2 - L_1^2\right) - 16R_1^2r_4^2\left(r_1^2 + R_1^2 - L_1^2\right)\right)\cos^2(\alpha) + \\
& 8R_1(\rho_3 - \rho_2)\left(r_1R_2\left((\rho_3 + \rho_2 - 2\rho_1)^2\right) + (r_1R_2 - r_4R_1)\left(2(x_p + D_1 - d_1)^2 + 6r_1^2 + 2R_1^2 - 2L_1^2\right)\right)\cos(\alpha)\sin(\alpha) - \\
& 2r_1R_1\left((\rho_3 - \rho_2)^4 + 4(\rho_3 - \rho_2)^3(\rho_2 - \rho_1) + 4(\rho_3 - \rho_2)^2\left((\rho_2 - \rho_1)^2 + (x_p + D_1 - d_1)^2 + 2r_1^2 + R_1^2 + R_2^2 - L_1^2\right) - \right. \\
& \left. 16R_2^2(\rho_3 - \rho_2)(\rho_1 - \rho_2) + 16\left(R_2^2\left((\rho_2 - \rho_1)^2 + r_1^2\right) + R_1^2r_4^2 - 2r_1R_1R_2r_4\right)\right)\cos(\alpha) - \\
& 4(\rho_3 - \rho_2)\left(R_2\left(R_1^2 + r_1^2\right)\left((\rho_3 - \rho_2)^2 + 4\left((\rho_2 - \rho_1)^2 + (\rho_3 - \rho_2)(\rho_2 - \rho_1)\right)\right) - \\
& 4r_1(r_4R_1 - r_1R_2)\left((x_p + D_1 - d_1)^2 + R_1^2 + r_1^2 - L_1^2\right)\right)\sin(\alpha) + \\
& 4(\rho_2 - \rho_1)^2(4R_2^2 + (\rho_3 - \rho_2)^2)(R_1^2 + r_1^2) + 4(\rho_2 - \rho_1)(\rho_3 - \rho_2)(4R_2^2 + (\rho_3 - \rho_2)^2)(R_1^2 + r_1^2) + \\
& (R_1^2 + r_1^2)(\rho_3 - \rho_2)^4 + 16(-R_1r_4 + r_1R_2)^2(r_1^2 + (x_p + D_1 - d_1)^2 - L_1^2 + R_1^2) + \\
& 4\left(r_1^2\left((x_p + D_1 - d_1)^2 + r_1^2 + R_1^2 + R_2^2 - L_1^2\right) + R_2^2R_1^2\right) = 0
\end{aligned} \qquad (19)$$

$$\begin{aligned}
&16R_2\Big(R_1(R_2r_1-R_1r_4)(\rho_3-\rho_2)^2+2r_4R_1^2\big((\rho_2-\rho_1)^2+(\rho_3-\rho_2)(\rho_2-\rho_1)\big)+2r_4(r_1R_2-r_4R_1)^2\Big)\cos^3(\alpha)+\\
&16R_2(\rho_3-\rho_2)\Big(R_1^2\big((\rho_2-\rho_1)^2+(\rho_3-\rho_2)(\rho_2-\rho_1)\big)+3R_1^2r_4^2+r_1^2R_2^2-4r_1R_1R_2r_4\Big)\cos^2(\alpha)\sin(\alpha)+\\
&4\Big(-R_1^2(\rho_2-\rho_1)(\rho_3-\rho_2)^3+\big(R_1^2\big((x_p+D_2-d_2)^2-(\rho_2-\rho_1)^2-r_4^2-L_3^2\big)+10r_1r_4R_1R_2-5r_1^2R_2^2\big)(\rho_3-\rho_2)^2-\\
&4R_1^2(r_4^2+R_2^2)\big((\rho_2-\rho_1)(\rho_3-\rho_2)-(\rho_2-\rho_1)^2\big)-4\big((x_p+D_2-d_2)^2+r_4^2+R_2^2-L_3^2\big)(r_1R_2-R_1r_4)^2\Big)\cos^2(\alpha)+\\
&4(\rho_3-\rho_2)\Big(R_1(3R_2r_1-r_4R_1)(\rho_3-\rho_2)^2+4(r_1R_2-R_1r4)\big(R_1\big((x_p+D_2-d_2)^2+r_4^2+R_2^2-L_3^2\big)+2r_1r_4R_2\big)\Big)\cos(\alpha)\sin(\alpha)+\\
&\Big(-32r_4R_2\big(R_1^2(\rho_2-\rho_1)^2+r_1^2R_2^2+r_4^2R_1^2-2r_1r_4R_1R_2\big)-32r_4R_1^2R_2(\rho_2-\rho_1)(\rho_3-\rho_2)+\\
&8\big(r_1R_1\big(-(x_p+D_2-d_2)^2-3R_2^2-r_4^2+L_3^2\big)+R_1^2R_2r_4-r_1^2R_2r_4\big)(\rho_3-\rho_2)^2-2r_1R_1(\rho_3-\rho_2)^4\Big)\cos(\alpha)+\\
&4\Big(2\big(2r_1(R_1r_4-r_1R_2)(x_p+D_2-d_2)^2-2R_2R_1^2(\rho_2-\rho_1)^2-2(r_1R_2-R_1r_4)(2r_1R_2^2+r_1r_4^2-r_4R_1R_2-r_1L_3^2)\big)(\rho_3-\rho_2)-\\
&4R_2R_1^2(\rho_2-\rho_1)(\rho_3-\rho_2)^2+(r_1r_4R_1-R_1^2R_2-2r_1^2R_2)(\rho_3-\rho_2)^3\Big)\sin(\alpha)+\\
&(r_1^2+R_1^2)(\rho_3-\rho_2)^4+4R_1^2(\rho_2-\rho_1)(\rho_3-\rho_2)^3+\\
&4\Big(R_1^2(\rho_2-\rho_1)^2+r_1^2(x_p+D_2-d_2)^2-r_1^2(L_3^2+r_4^2+6R_2^2)+R_1^2R_2^2-6r_1r_4R_1R_2+2r_4^2R_1^2\Big)(\rho_3-\rho_2)^2+\\
&16R_1^2(r_4^2+R_2^2)\big((\rho_2-\rho_1)(\rho_3-\rho_2)+(\rho_2-\rho_1)^2\big)+16\big(R_2^2+(x_p+D_2-d_2)^2-L_3^2+r_4^2\big)(r_1R_2-r_4R_1)^2=0
\end{aligned}$$
(20)

## 7. Acknowledgments

This work has been partially funded by the European projects NEXT, acronyms for "Next Generation of Productions Systems", Project no° IP 011815. The authors would like to thank the Fatronik society, which permitted us to use the CAD drawing of the Machine VERNE what allowed us to present well the machine. The authors would also like to thank Professor Wisama KHALIL for his useful remarks that helped us accomplishing this work.